\pdfoutput=1
\documentclass[journal]{IEEEtran}
\usepackage{graphicx}
\usepackage{placeins}           % Allows the use of \FloatBarrier, which means all floats must be placed before continuing
\usepackage[export]{adjustbox}  % Allows for scaling, then max width/height
\usepackage{amsmath}
\usepackage{amssymb}            % for \mathbb (like R meaning all real numbers, etc.)
\usepackage{cite}

\usepackage{xcolor}             % for showing changes

\title{Mesh Graph Neural Network Framework for Accelerating Finite Element Simulation for Arbitrary Geometries}

\author{%
\IEEEauthorblockN{Josiah D. Kunz\IEEEauthorrefmark{1} and Kamal Choudhary\IEEEauthorrefmark{2}}\\
\IEEEauthorblockA{\IEEEauthorrefmark{1}Illinois College,\\
Jacksonville, IL, USA\\
Email: \texttt{josiah.kunz@ic.edu}}\\
\and
\IEEEauthorblockA{\IEEEauthorrefmark{2}Department of Materials Science and Engineering, Whiting School of Engineering, The Johns Hopkins University, Baltimore, MD 21218, USA\\
Email: \texttt{drkamal@jhu.edu}}%
}

\begin{document}
\maketitle

% ------------------------------------------------------------------------------------------------------------------

\begin{abstract}
    Finite element analysis (FEA) is essential for structural design but remains computationally expensive, particularly when evaluating multiple design iterations or load scenarios. Machine learning surrogate models offer a promising alternative, yet most approaches struggle with a critical limitation: generalizing across varying geometries. This work presents a mesh graph network (MGN) for predicting von~Mises stress fields in 2D structural components with arbitrary hole geometries. Unlike traditional machine learning approaches that use absolute node coordinates as features, the proposed model builds on existing MGN frameworks that encode node types (e.g., fixed boundary, free surface, hole edge), relative edge features (distance between neighbors), and global features (applied load). This architecture is inherently translation- and rotation-invariant, enabling generalization to unseen geometries without retraining. The MGN was trained on 11 plate geometries under 20 load conditions and evaluated on 7 unseen geometries and 3 unseen loads. In the most favorable case, the model achieves \boldmath$R^2 \geq 0.97$\unboldmath{} on an unseen geometry and unseen load, compared to \boldmath$R^2 \approx 0.01$--$0.86$\unboldmath{} for conventional models (Random Forest \cite{breiman2001randomforest}, Gradient Boosting \cite{friedman2001gboost}, K-Nearest Neighbors \cite{cover1967nearestneighbor}) trained on identical data. However, even in less favorable cases, the MGN model still outperforms conventional models. This work extends the mesh-based simulation framework of Pfaff et al.\ \cite{clothmgn} to structural mechanics, demonstrating that graph neural networks can serve as efficient surrogates for finite element analysis across varying geometries.
\end{abstract}

% ------------------------------------------------------------------------------------------------------------------

\section{Introduction}

\IEEEPARstart{F}{inite} element analysis (FEA) is the standard approach for predicting stress, strain, and deformation in structural components \cite{bathe2006, hughes2012, zienkiewicz2013}. However, FEA simulations are computationally expensive, often requiring minutes to hours per evaluation depending on mesh resolution and problem complexity \cite{alizadeh2020, jiang2020surrogate, liu2022eightyears}. This computational cost limits the use of FEA in applications requiring rapid iteration, such as design optimization, uncertainty quantification, and real-time digital twins.

Machine learning (ML) surrogates offer a promising alternative, providing near-instantaneous predictions once trained \cite{karniadakis2021, raissi2019, gu2024statistical}. Traditional ML approaches for structural mechanics typically use node coordinates and loading conditions as input features to predict field quantities such as stress or displacement \cite{liang2018deep, nie2020stress, greve2022surrogate}. While effective for interpolation within a fixed geometry, these methods fail to generalize across different geometries because they memorize absolute positions rather than learning underlying physical relationships.

Graph neural networks (GNNs) address this limitation by aggregating information from neighboring nodes. By including this connectivity, GNNs have demonstrated success in materials science for predicting properties of crystals and molecules \cite{xie2018, choudhary2021}, as well as grain-scale mechanical behavior \cite{pagan2022}.

Recent work has also applied GNNs to structural mechanics problems, including stress and strain field prediction in composites and metamaterials \cite{maurizi2022}, plate-with-hole problems \cite{gulakala2022, gulakala2024, wu2024}, and physics-informed approaches for general engineering structures \cite{hou2026}. However, these approaches still include absolute node coordinates (or in the case of \cite{gladstone2024}, transformed coordinates) as input features. While effective for interpolation, coordinate-based representations limit generalization to new geometries and orientations. 

Pfaff et al.\ \cite{clothmgn} introduced mesh-based graph networks (MGNs) that encode node types into vector representations instead of quantifying nodes by their positions. Using relative edge features, information propagates through message passing, enabling generalization across different mesh configurations \cite{battaglia2018relational}. For simulating complex physical systems, recent work has shown effectiveness in cloth dynamics \cite{clothmgn}, fluid flow \cite{sanchez2020}, and particle-based materials \cite{sanchez2020}. W\"{u}rth et al.\ \cite{wuerth2024pimgn} extended MGNs with physics-informed training for thermal simulations, demonstrating that models trained on small meshes can generalize to much larger domains when using relative positional encoding rather than absolute coordinates. However, the application of MGNs to structural mechanics problems with varying geometries remains underexplored.

This work makes three contributions. First, an MGN framework is presented for predicting von~Mises stress fields in 2D structural components with arbitrary hole geometries. The proposed architecture achieves translation and rotation invariance through node-type embeddings, relative edge features, and global load encoding, which eliminates the dependence on absolute coordinates. Generalization to unseen geometries is demonstrated, achieving $R^2 \geq 0.97$ on geometrically similar test cases. Second, we provide a systematic comparison against traditional ML models (Random Forest, Gradient Boosting, K-Nearest Neighbors), showing consistent improvement on unseen geometries and loads. Finally, we analyze failure modes relevant to this model, identifying geometric dissimilarity as a key factor limiting generalization.

The geometry-agnostic nature of this framework also points toward broader materials science applications. Stress fields around voids, pores, or inclusions in polycrystalline microstructures span length scales that are inaccessible to atomistic methods, such as molecular dynamics or density functional theory, where the required number of particles or simulation cells grows prohibitively large \cite{jia2020}. At these scales, continuum FEA remains necessary. For example, Hartmaier \cite{hartmaier2023} demonstrated a CNN surrogate for predicting von~Mises stress fields in viscoplastic polycrystalline microstructures, achieving $\sim$500$\times$ speedup over direct numerical solutions. A geometry-generalizable MGN could extend such approaches to arbitrary pore and inclusion geometries without retraining.

\begin{figure*}[ht!]
    \centering
    \includegraphics[scale=1, max width=\linewidth, max height=0.9\textheight]{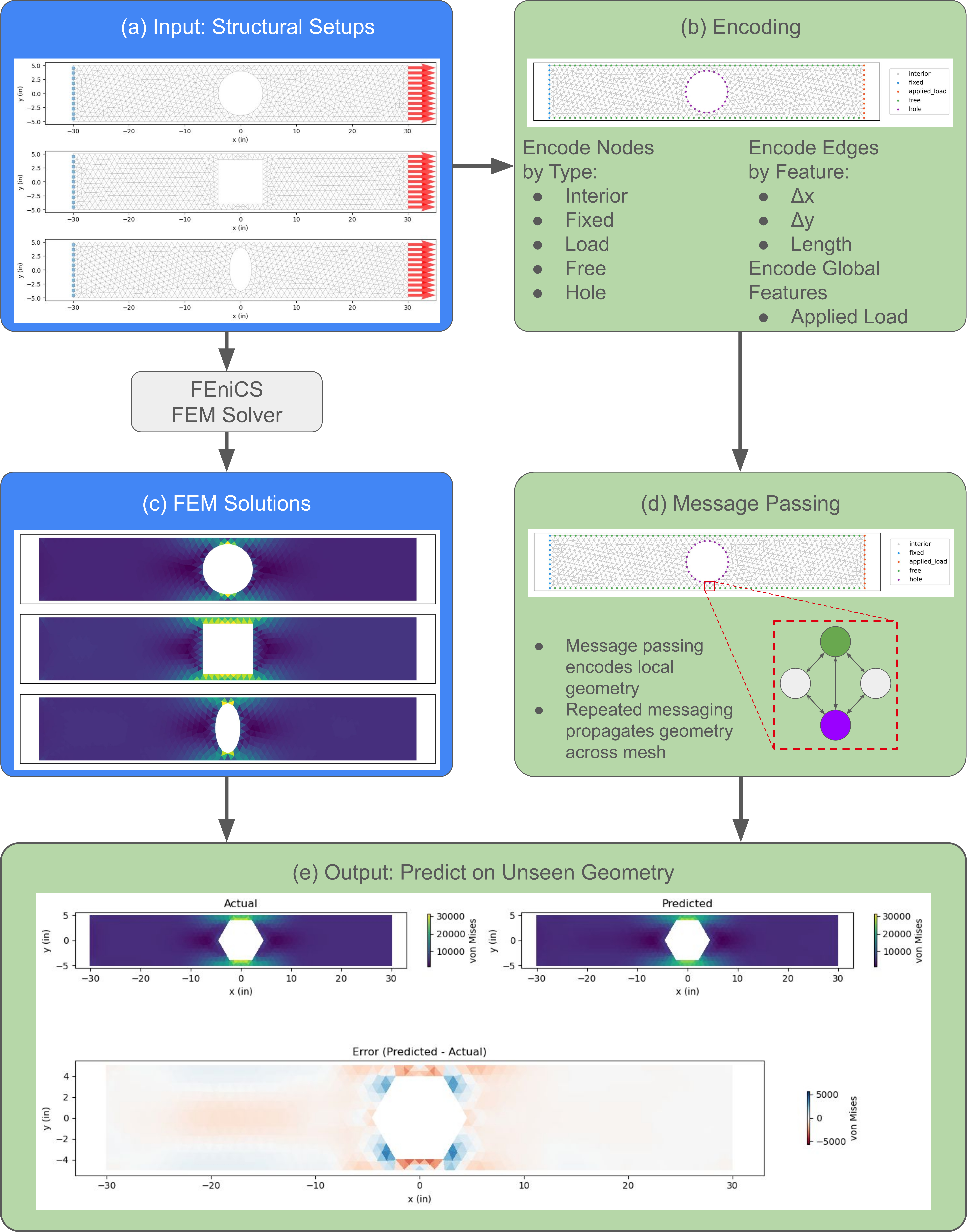}
    \caption{Overview schematic. (a) Varying geometries and boundary conditions serve as inputs. (b) The mesh graph network (MGN) encodes each node by type, mapping to a learned embedding vector. Edge features $(\Delta x, \Delta y, \ell)$ and the global feature (applied load) are encoded via separate MLPs. (c) FEM solutions (FEniCS) provide training targets. (d) The training loop includes multiple message-passing layers to propagate local geometry globally. (e) The trained MGN predicts stress on an unseen geometry and is compared to the FEniCS ground truth.}
    \label{fig:schematic}
\end{figure*}

% ------------------------------------------------------------------------------------------------------------------

\section{Results}

\subsection{Seen Geometries with Unseen Loads}

The MGN was first evaluated on training geometries under loads not seen during training (800, 5000, and 12000~psi). Results are summarized in Table~\ref{tab:seen-geom-unseen-load} and visualized in Figure~\ref{fig:unseen-loads}.

For most geometries, the model achieves $R^2 > 0.99$ across all unseen loads, demonstrating strong interpolation and extrapolation capability within the training geometry distribution. However, two geometries show reduced performance: the 1" hole (geometry i) and the no-hole plate (geometry j).

These geometries produce stress distributions underrepresented in the training data. The 1" hole creates highly localized stress concentrations with few hole-boundary nodes, while the no-hole plate produces nearly uniform stress fields with no hole-boundary nodes at all. Both patterns differ substantially from the dominant training examples, which feature prominent hole boundaries. The learned embeddings for \texttt{hole}-type nodes likely carry significant weight in the model's predictions, causing performance to degrade when such nodes are sparse or absent.

\subsection{Unseen Geometries}

The primary test of generalization is performance on geometries not seen during training (Figure~\ref{fig:unseen_geom_grid}). Results varied substantially depending on geometric similarity to training data.

The hexagonal hole ($R^2 = 0.97$) performed best, being geometrically similar to circular and square holes in training data. The triangular hole ($R^2 = 0.71$) showed moderate performance despite sharp 60$^\circ{}$ corners not present in training, which contained only the 90$^\circ{}$ corners of the square hole. Performance degraded substantially for the figure-8 hole ($R^2 = 0.32$), a novel geometry with two connected lobes and dense mesh refinement around the sharp corners at the cleavage, resulting in a high concentration of \texttt{interior}-type nodes in stress concentration regions. The 8'' J-shaped hole ($R^2 < 0$) failed entirely, presenting a highly dissimilar, asymmetric stress pattern with similar mesh refinement issues near the hook.

These results indicate that the MGN generalizes well to geometries producing similar stress patterns but struggles with geometrically novel configurations. Two potential remedies are identified:

\begin{enumerate}
    \item \textbf{Sharp corners}: Introducing a dedicated \texttt{corner} node type could help the model distinguish stress concentration regions from smooth boundaries.
    \item \textbf{Dense interior regions}: Information propagation through regions with many \texttt{interior}-type nodes could be improved by either increasing the number of message-passing layers $L$ or by employing multi-hop aggregation where each layer gathers information from neighbors 2 or more hops away.
\end{enumerate}

\subsection{Comparison with Traditional ML Models}

To contextualize MGN performance, traditional ML models were trained on identical data using absolute coordinates $[x, y, \text{load}]$ as features (Figure~\ref{fig:ml-comparison}). On the unseen hexagonal geometry, the MGN achieved $R^2 = 0.97$, compared to $R^2 = 0.62$ for Random Forest, $R^2 = 0.45$ for Gradient Boosting, and $R^2 = 0.16$ for K-Nearest Neighbors.

Traditional models fail to generalize because they memorize coordinate-to-stress mappings rather than learning geometric relationships. The MGN's use of relative edge features and node-type embeddings enables transfer to new geometries.

\subsection{Error Distribution by Node Type}

Figure~\ref{fig:hex-by-node-type} shows prediction accuracy stratified by node type for the hexagonal hole at 5000~psi. The largest prediction errors occur at hole boundary nodes, where stress concentrations are highest and gradients are steepest. Interior and exterior boundary nodes show smaller errors consistent with their smoother stress distributions.

This further supports the observation that the model relies heavily on \texttt{hole}-type nodes. Earlier, performance degraded when hole-boundary nodes were sparse (1" hole) or absent (no-hole plate). Here, even when hole boundaries are well-represented, the model's errors concentrate at these nodes. Together, these findings suggest that \texttt{hole}-type nodes dominate the learned representations; that is, the model depends on them for accurate predictions, and prediction quality is most sensitive to getting them right.

\begin{figure*}[p]
    \centering
    \includegraphics[scale=2, max width=\linewidth, max height=0.9\textheight]{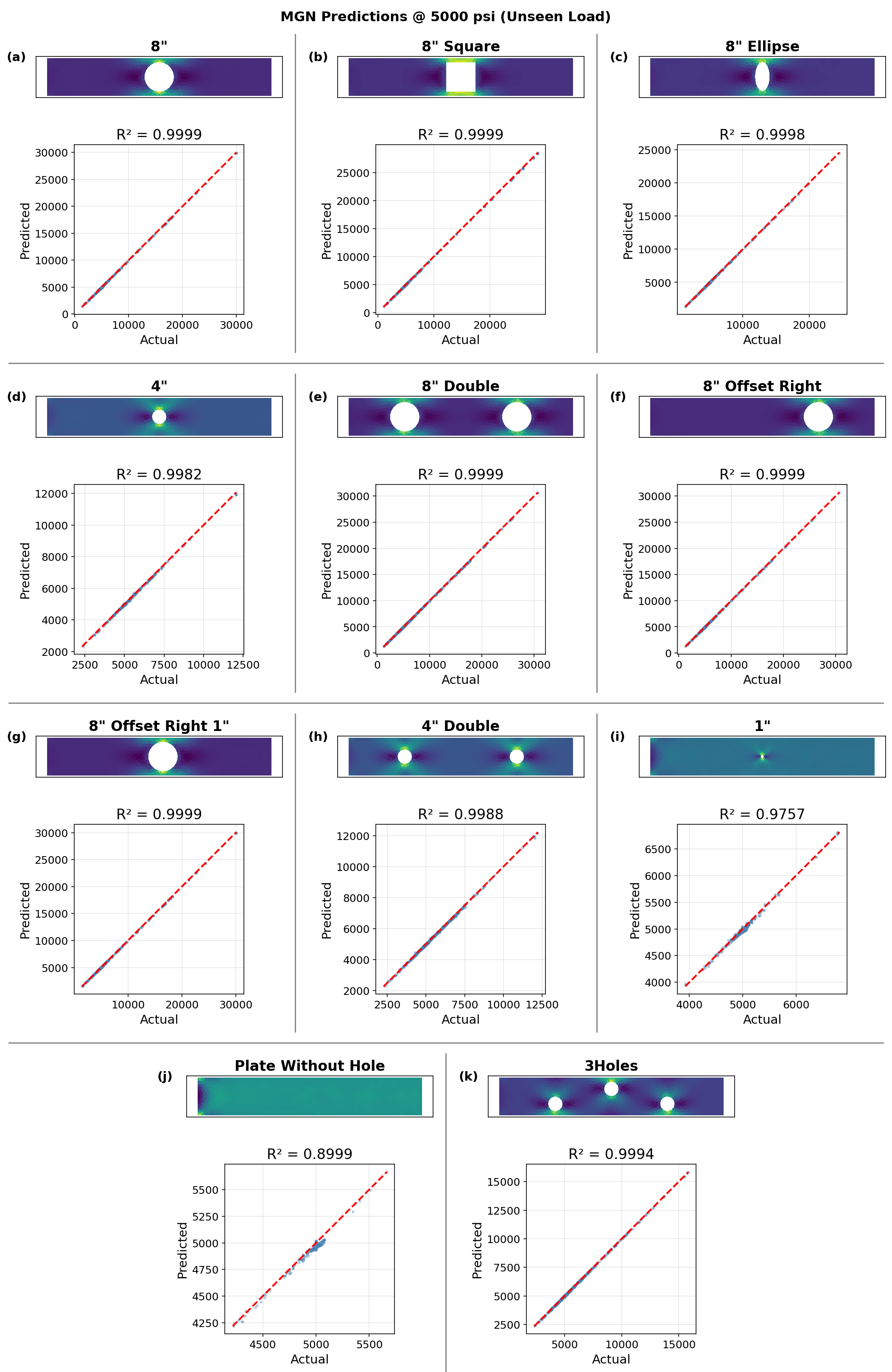}
    \caption{Mesh graph network (MGN) predictions as (top) von Mises field values and (bottom) scatterplots compared against ground truth for seen geometries under unseen loads. MGN was trained on 20 loads per geometry with a range of $(1000, 4500)$ and $(5500, 10000)$~psi.}
    \label{fig:unseen-loads}
\end{figure*}

\begin{figure*}[p]
    \centering
    \includegraphics[scale=2, max width=\linewidth, max height=0.9\textheight]{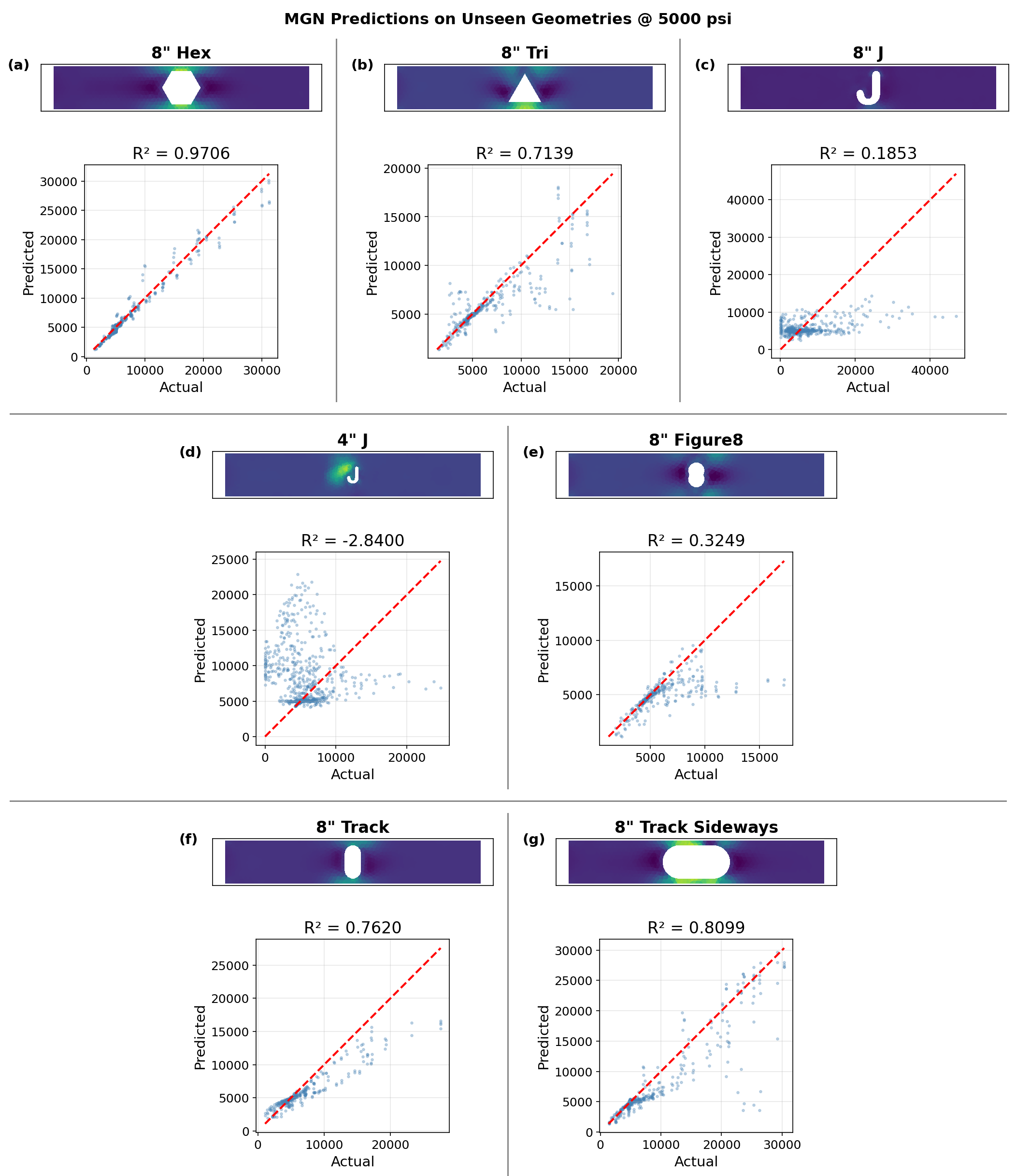}
    \caption{Mesh graph network (MGN) predictions as (top) von Mises field values and (bottom) scatterplots compared against ground truth for unseen geometries under unseen loads.}
    \label{fig:unseen_geom_grid}
\end{figure*}

\begin{figure*}[p]
    \centering
    \includegraphics[scale=1, max width=\linewidth, max height=0.9\textheight]{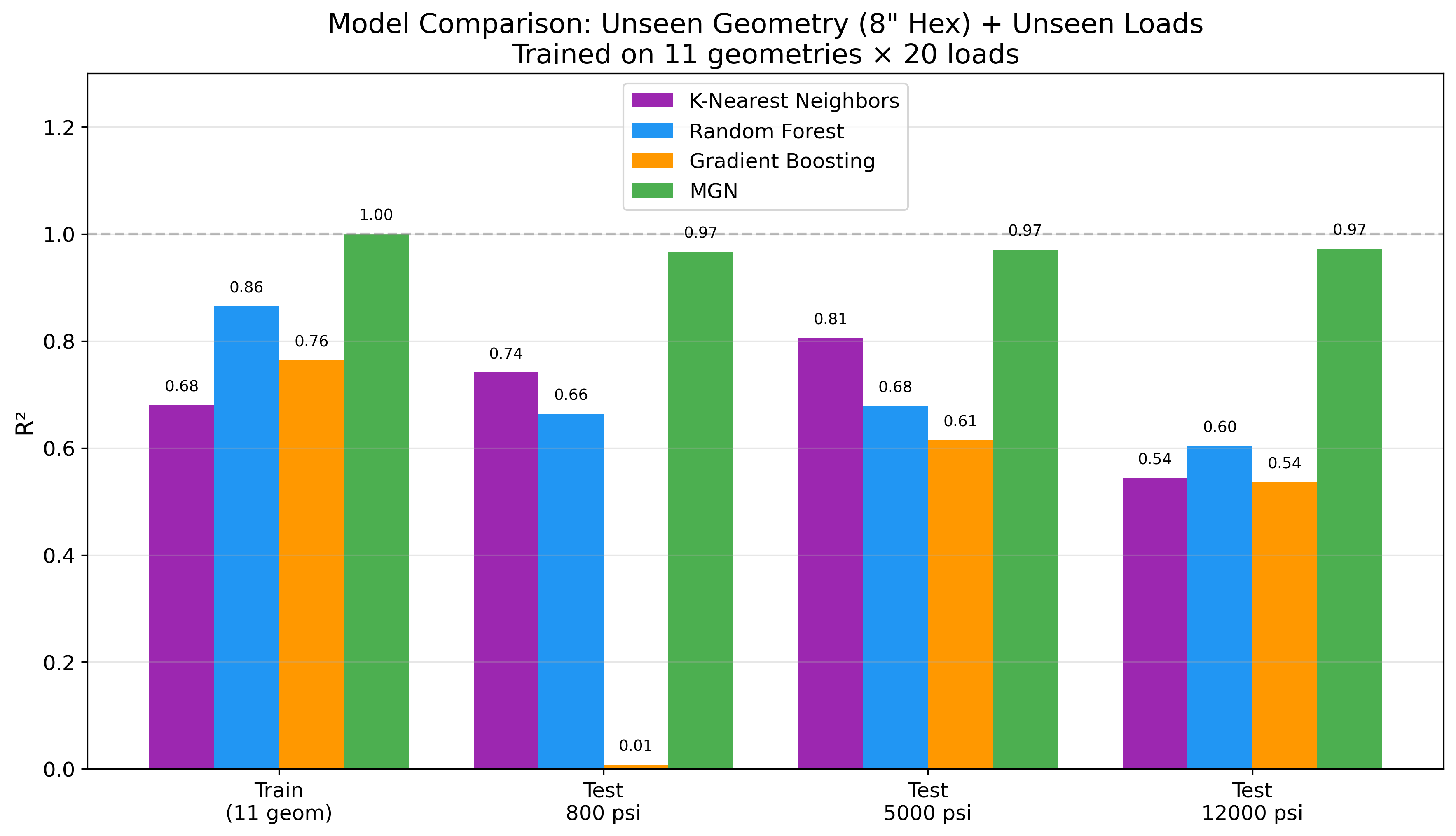}
    \caption{Mesh graph network (MGN) compared to traditional \texttt{sklearn} models, first on the trained data, then on an unseen geometry (8" hexagonal hole) with unseen loads (800, 5000, and 12000~psi). All models were trained on identical data: the 11~geometries found in Figure~\ref{fig:geometries} with 20 different applied loads $\in [1000, 4500] \cup [5500, 10000]$~psi. Traditional models use absolute coordinates and load $[x, y, \text{load}]$ as features, predicting von Mises stress per node. While this node-level approach allows evaluation on different mesh sizes, it fails to generalize to new geometries because it memorizes positions rather than learning geometric relationships. MGN maintains R$^2 \geq 0.97$ by using relative edge features and node-type embeddings.}
    \label{fig:ml-comparison}
\end{figure*}

\begin{figure*}[p]
    \centering
    \includegraphics[scale=1, max width=\linewidth, max height=0.9\textheight]{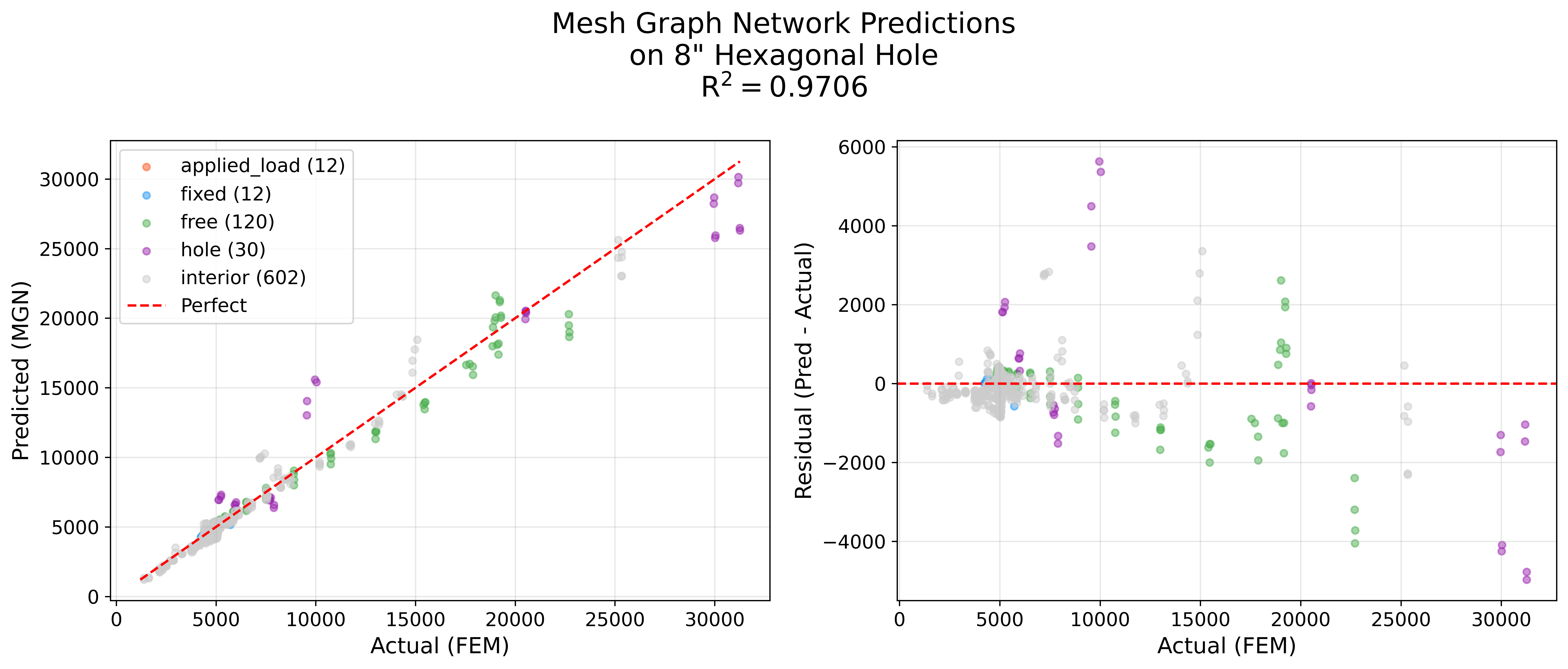}
    \caption{Mesh graph network (MGN) prediction accuracy by node type for the 8" hexagonal hole (unseen geometry) at 5000~psi (unseen load). Points are colored by their classification: interior nodes (gray), fixed boundary (blue), free boundary (green), hole boundary (purple), and applied load (orange). The model performs well across all node types, with the largest deviations occurring at hole boundaries where stress concentrations are highest.}
    \label{fig:hex-by-node-type}
\end{figure*}

% ------------------------------------------------------------------------------------------------------------------

\section{Discussion}

\subsection{Why MGN Generalizes}

The MGN's ability to generalize to unseen geometries stems from three design choices:

\begin{enumerate}
    \item \textbf{Node-type embeddings}: Rather than encoding absolute position, nodes are classified by their role (fixed, free, hole, etc.), which transfers across geometries.
    \item \textbf{Relative edge features}: Edge features $[\Delta x, \Delta y, \ell]$ encode local geometry without reference to a global coordinate system, providing translation and rotation invariance.
    \item \textbf{Global conditioning}: The applied load is encoded separately and broadcast to all nodes, decoupling loading conditions from geometry.
\end{enumerate}

\subsection{Limitations and Future Work}

The current implementation handles planar geometries. Extension to 3D follows naturally: edge features generalize to three dimensions, and similar graph-based approaches have been demonstrated for 3D structures \cite{clothmgn, maurizi2022, gladstone2024}. However, 3D meshes would substantially increase computational cost due to larger node counts.

Beyond geometric dimension, several other extensions remain unexplored. Only linear elastic material behavior was considered; plasticity and other nonlinearities would require history-dependent models, such as path-dependent architectures \cite{mozaffar2019deep} or mixture density networks \cite{luo2025uncertainty}. Similarly, only uniaxial tension was studied, and generalization to multi-axial loading, combined loading (e.g., tension plus bending), or time-varying loads remains untested.

The model also has resolution constraints. Training used a fixed mesh size of 25~mm, and the model does not generalize to coarser or finer discretizations. However, this is an active area of research \cite{gladstone2024, franco2023}. Relatedly, the number of message-passing layers depends on mesh density. For this work, 20 layers were used, but denser meshes may require more layers for information about the hole to propagate across the entire mesh. Improvements might include edge-augmented GNNs or multi-GNNs, both used in \cite{gladstone2024}.

Finally, while inference is fast (under one second per geometry), training required 10,000 epochs over 220 graphs. For rapid prototyping or active learning scenarios, this training cost may be a bottleneck.

% ------------------------------------------------------------------------------------------------------------------

\section{Methods}

\subsection{Problem Formulation}

A 2D plate with up to three holes subjected to uniaxial tension is considered. The goal is to predict the von~Mises stress field $\sigma_{\text{vM}}(\mathbf{x})$ \cite{timoshenko1951theory} at each mesh node $\mathbf{x}$ given the geometry and boundary conditions. Traditional finite element analysis solves this problem by discretizing the domain into elements and solving the resulting system of equations, which can be computationally expensive for complex geometries or iterative design processes.

The problem is formulated as a supervised learning task: given an attributed graph $G = (V, E, \mathbf{u})$ where $V$ are nodes, $E$ are edges, and $\mathbf{u} \in \mathbb{R}^k$ are global attributes, predict the von~Mises stress at each node. The key challenge is designing an architecture that generalizes across different geometries without retraining.

\subsection{Graph Representation}

The finite element mesh is converted to a graph $G = (V, E, \mathbf{u})$ where nodes $V$ correspond to mesh vertices, edges $E$ correspond to mesh connectivity, and $\mathbf{u}$ represents global attributes. Following Pfaff et al.\ \cite{clothmgn}, relative features are used rather than absolute coordinates to achieve translation and rotation invariance.

\subsubsection{Node Features}

Each node $i$ is assigned a categorical type $t_i \in \{$\texttt{interior}, \texttt{fixed}, \texttt{free}, \texttt{hole}, \texttt{applied\_load}$\}$ based on its position in the mesh and the applied boundary conditions:

\begin{table}[ht]
\centering
\caption{Node type classifications.}
\label{tab:node-types}
\begin{tabular}{ll}
\hline
\textbf{Node Type} & \textbf{Description} \\
\hline
\texttt{fixed} & Boundary nodes on the constrained edge \\
\texttt{applied\_load} & Boundary nodes where the load is applied \\
\texttt{hole} & Nodes on interior hole boundaries \\
\texttt{free} & Remaining exterior boundary nodes \\
\texttt{interior} & Remaining nodes \\
\hline
\end{tabular}
\end{table}

To distinguish holes from the exterior boundary, all closed boundary loops are identified, and then the loop enclosing the largest area is classified as the exterior perimeter. All other loops are classified as holes.

Node types are embedded into a learned vector representation $\text{Embed}(t_i) \in \mathbb{R}^{d_e}$, where $d_e$ is the embedding dimension. This embedding is then projected to the hidden dimension $d_h$ via an MLP:
\begin{equation}
    \mathbf{h}_i^{(0)} = \text{MLP}_{\text{node}}\left( \text{Embed}(t_i) \right) \in \mathbb{R}^{d_h}
\end{equation}
For example, the node type \texttt{interior} may be embedded as a learned 16-valued vector $[0.23, -0.15, ..., 0.42]$. 

\subsubsection{Edge Features}

Each edge $(i, j) \in E$ is characterized by three features:
\begin{equation}
    \mathbf{e}_{ij} = \left[ \Delta x_{ij}, \Delta y_{ij}, \ell_{ij} \right]
\end{equation}
where $\Delta x_{ij} = x_j - x_i$ and $\Delta y_{ij} = y_j - y_i$ are the relative displacements, and $\ell_{ij} = \sqrt{\Delta x_{ij}^2 + \Delta y_{ij}^2}$ is the Euclidean distance. These relative features encode local geometry without reference to a global coordinate system. Edge features are directly projected to the hidden dimension (without embedding) via a dedicated MLP:
\begin{equation}
    \mathbf{e}_{ij}^{(0)} = \text{MLP}_{\text{edge}}\left( \mathbf{e}_{ij} \right) \in \mathbb{R}^{d_h}
\end{equation}

\subsubsection{Global Features}

The applied load magnitude $p$ is provided as a global feature, broadcast to all nodes after encoding through yet another separate MLP:
\begin{equation}
    \mathbf{u}^{(0)} = \text{MLP}_{\text{global}}(p) \in \mathbb{R}^{d_h}
\end{equation}
This allows the network to condition predictions on loading conditions while maintaining geometric invariance. 

\subsubsection{Processor}

The processor propagates information across the graph through $L$ message-passing layers (default: 20). Each node maintains a hidden state $\mathbf{h}_i^{(\ell)} \in \mathbb{R}^{d_h}$ that is iteratively updated by aggregating information from neighboring nodes.

At each layer $\ell$, a message $\mathbf{m}_{ij}^{(\ell)}$ is computed for each edge, representing information sent from node $j$ to node $i$:
\begin{equation}
    \mathbf{m}_{ij}^{(\ell)} = \text{MLP}_{\text{msg}}^{(\ell)}\left( \mathbf{h}_j^{(\ell)}, \mathbf{e}_{ij}^{(0)} \right)
\end{equation}
where $\mathbf{h}_j^{(\ell)}$ is the hidden state of the neighboring node and $\mathbf{e}_{ij}^{(0)}$ is the encoded edge feature.

Messages from all neighbors $\mathcal{N}(i)$ are summed and used to update the node's hidden state:
\begin{equation}
    \mathbf{h}_i^{(\ell+1)} = \text{MLP}_{\text{update}}^{(\ell)}\left( \mathbf{h}_i^{(\ell)}, \sum_{j \in \mathcal{N}(i)} \mathbf{m}_{ij}^{(\ell)}, \mathbf{u}^{(0)} \right)
\end{equation}

This process repeats for $L$ layers, allowing information to propagate across $L$ hops in the graph. For instance, $L=1$ limits each node's receptive field to its immediate neighbors, whereas $L=20$ (used in this work) allows information to propagate across the entire domain.

\subsubsection{Output Layer}

The final node embeddings are decoded to predict von~Mises stress:
\begin{equation}
    \hat{\sigma}_{\text{vM},i} = \text{MLP}_{\text{decode}}\left( \mathbf{h}_i^{(L)} \right)
\end{equation}

\subsection{Dataset Generation}

\subsubsection{Geometry}

\begin{figure*}[ht!]
    \centering
    \includegraphics[scale=2, max width=\linewidth, max height=0.9\textheight]{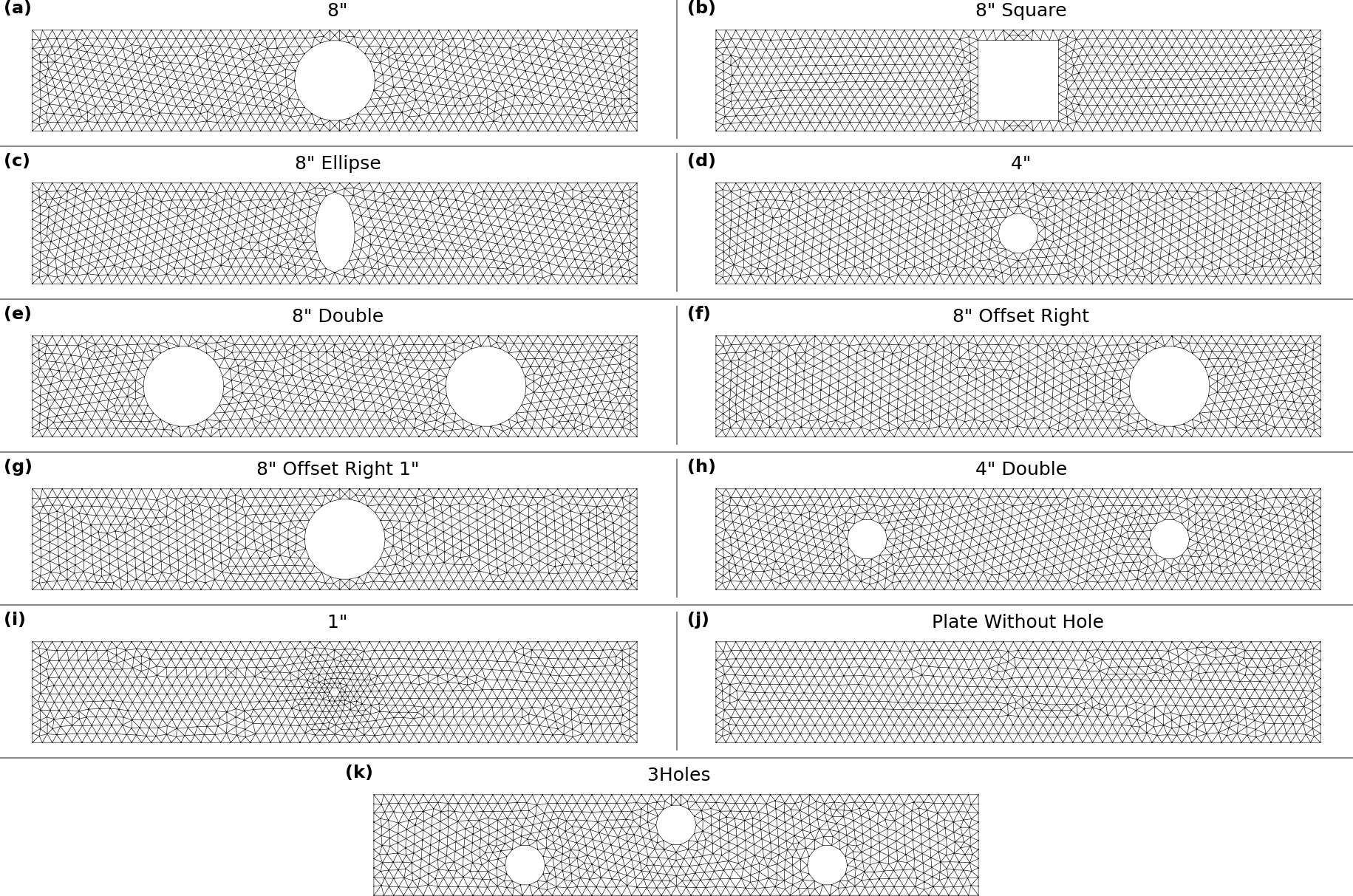}
    \caption{Meshes used to train the mesh graph network (MGN). The plate dimensions were 60"$\times$10". The STEP files were created in Creo and then converted to a mesh via gmsh. (a) 8" circle (b) 8" square (c) 8"$\times$4" ellipse (d) 4" circle (e) two 8" circles at $x=\pm15$" (f) one 8" circle offset to the right by 15" (g) one 8" circle offset to the right by 1" (h) two 4" circles at $x=\pm15$" (i) one 1" circle (j) no holes (k) three holes.}
    \label{fig:geometries}
\end{figure*}

Training data was generated from 11 plate geometries (Figure~\ref{fig:geometries}), all with outer dimensions of $60'' \times 10''$. Geometries included circular holes of varying sizes (1'', 4'', 8'' diameter), non-circular holes (square, elliptical), multiple holes (double circles, triple holes), offset hole positions, and a plate without holes.

CAD models were created in PTC Creo and exported as STEP files. Meshes were generated using Gmsh \cite{gmsh} with triangular elements and a target element size of 25~mm.

\subsubsection{Loading Conditions}

Each geometry was simulated under 20 uniaxial tensile loads spanning $[1000, 4500] \cup [5500, 10000]$~psi, applied as traction on the right edge. The left edge was fixed (zero displacement). The gap in load range was intentional, reserving 5000~psi for interpolation testing.

\subsubsection{FEM Simulation}

Simulations were performed using FEniCS \cite{fenics} with linear elastic material properties (steel: $E = 200$~GPa, $\nu = 0.3$). Plane stress conditions were assumed. Von~Mises stress was computed from the stress tensor at each node.

The complete training set comprised $11 \times 20 = 220$ simulations, yielding approximately 180,000 node-level samples.

\subsection{Training}

The model was trained to minimize mean squared error between predicted and FEM-computed von~Mises stress:
\begin{equation}
    \mathcal{L} = \frac{1}{N} \sum_{i=1}^{N} \left( \hat{\sigma}_{\text{vM},i} - \sigma_{\text{vM},i}^{\text{FEM}} \right)^2
\end{equation}

Training was performed using the Adam optimizer \cite{adam} with a learning rate of $10^{-5}$ for 10,000 epochs. The model was implemented in PyTorch \cite{pytorch} with PyTorch Geometric \cite{pyg} for graph operations.

Hyperparameters were set as follows: hidden dimension $d_h = 64$, embedding dimension $d_e = 16$, number of message-passing layers $L = 20$, and batch size of 1 graph per iteration (with all nodes in the graph processed simultaneously). As previously mentioned, the training set comprised 11 geometries $\times$ 20 load cases $=$ 220 simulations.

\subsection{Evaluation}

Model performance was evaluated using the coefficient of determination:
\begin{equation}
    R^2 = 1 - \frac{\sum_{i}(\sigma_{\text{vM},i}^{\text{FEM}} - \hat{\sigma}_{\text{vM},i})^2}{\sum_{i}(\sigma_{\text{vM},i}^{\text{FEM}} - \bar{\sigma}_{\text{vM}}^{\text{FEM}})^2}
\end{equation}

Generalization was tested on:
\begin{enumerate}
    \item \textbf{Unseen loads}: Loads outside the training range (800, 5000, 12000~psi) on training geometries (Figure~\ref{fig:unseen-loads}, Table~\ref{tab:seen-geom-unseen-load})
    \item \textbf{Unseen geometries}: Seven geometries not seen during training: hexagonal, triangular, 8"~J, 4"~J, figure 8, track, and sideways track (Figure~\ref{fig:unseen_geom_grid})
\end{enumerate}

Additionally, the MGN was compared against traditional ML models (Random Forest \cite{breiman2001randomforest}, Gradient Boosting \cite{friedman2001gboost}, K-Nearest Neighbors \cite{cover1967nearestneighbor}) trained on identical data using absolute coordinates as features. Prediction accuracy was also analyzed by node type to identify where errors concentrate (Figure~\ref{fig:hex-by-node-type}).

\FloatBarrier
% ------------------------------------------------------------------------------------------------------------------

\begin{table}[ht!]
\caption{$R^2$ values for mesh graph network (MGN) predictions on seen geometries with varying unseen loads. Note: geometries (i) and (j) show reduced performance at extreme loads, likely due to underrepresentation of similar stress distributions in training data.}
\label{tab:seen-geom-unseen-load}
\centering
\begin{tabular}{l|ccc}
\hline
\textbf{Geometry} & \textbf{800 psi} & \textbf{5000 psi} & \textbf{12000 psi} \\ \hline
(a) 8" & 0.9944 & 0.9999 & 0.9993 \\
(b) 8" Square & 0.9961 & 0.9999 & 0.9995 \\
(c) 8" Ellipse & 0.9930 & 0.9998 & 0.9989 \\
(d) 4" & 0.9733 & 0.9982 & 0.9951 \\
(e) 8" Double & 0.9950 & 0.9999 & 0.9992 \\
(f) 8" Offset 15" & 0.9951 & 0.9999 & 0.9992 \\
(g) 8" Offset 1" & 0.9948 & 0.9999 & 0.9994 \\
(h) 4" Double & 0.9795 & 0.9988 & 0.9975 \\
(i) 1" & 0.5829 & 0.9757 & 0.8816 \\
(j) No Hole & -0.0088 & 0.8999 & 0.7142 \\
(k) 3 Holes & 0.9826 & 0.9994 & 0.9981 \\ \hline
\end{tabular}
\end{table}

% ------------------------------------------------------------------------------------------------------------------

\section{Conclusion}

A mesh graph network framework was presented for predicting von~Mises stress fields in 2D plates with arbitrary hole geometries. By encoding node types, relative edge features, and global load parameters (rather than absolute coordinates), the model achieves generalization to unseen geometries.

On geometries similar to training data, the MGN achieves $R^2 \geq 0.97$, substantially outperforming traditional ML models that rely on coordinate-based features. Performance degrades for dissimilar geometries, but could be improved by several methods, including simply diversifying the training.

This work demonstrates that graph neural networks can serve as efficient, geometry-agnostic surrogates for finite element analysis, with potential applications in design optimization, uncertainty quantification, and real-time structural assessment.

% ------------------------------------------------------------------------------------------------------------------

\section*{Acknowledgments}

Computational resources were provided by Johns Hopkins University and Illinois College.

The authors acknowledge the use of Claude (Anthropic) for assistance with drafting, editing, and refining the manuscript text, as well as debugging code and serving as a sounding board. All technical content, experimental design, and scientific conclusions are the authors' own.

% ------------------------------------------------------------------------------------------------------------------

\section*{Code and Data Availability}

The MGN implementation, training scripts, data generation, and evaluation code are available at:
\texttt{https://github.com/Josiah-Kunz/MGN-Public}

The \texttt{meshgraphnet} Python package is also available at:
\texttt{https://pypi.org/project/meshgraphnet/}

% ------------------------------------------------------------------------------------------------------------------

\section*{Author Contributions}
J.K. designed the geometry STEP files, wrote and executed the code, and designed the algorithms used. K.C. made decisions on the project's direction and contributed to the code architecture. All authors reviewed and edited the manuscript.

% ------------------------------------------------------------------------------------------------------------------

\section*{Competing Interests}
The authors have neither financial nor non-financial competing interests.

% ------------------------------------------------------------------------------------------------------------------

\FloatBarrier
\bibliographystyle{IEEEtran}
\bibliography{references}

\end{document}